\documentclass{article}
\usepackage{spconf,amsmath,graphicx}
\usepackage{times}
\usepackage{epsfig}
\usepackage{amssymb,mathtools}
\usepackage[dvipsnames]{xcolor}

\title{INDOOR DENSE DEPTH MAP AT DRONE HOVERING}
\name{Arindam Saha, Soumyadip Maity, Brojeshwar Bhowmick}
\address{Embedded Systems and Robotics, TCS Research \& Innovation, Kolkata, India}
\begin{document}
%
\maketitle
\begin{abstract}
Autonomous Micro Aerial Vehicles (MAVs) gained tremendous attention in recent years. Autonomous flight in indoor requires a dense depth map for navigable space detection which is the fundamental component for autonomous navigation. In this paper, we address the problem of reconstructing dense depth while a drone is hovering (small camera motion) in indoor scenes using already estimated cameras and sparse point cloud obtained from a vSLAM. We start by segmenting the scene based on sudden depth variation using sparse 3D points and introduce a patch-based local plane fitting via energy minimization which combines photometric consistency and co-planarity with neighbouring patches. The method also combines a plane sweep technique for image segments having almost no sparse point for initialization. Experiments show, the proposed method produces better depth for indoor in artificial lighting condition, low-textured environment compared to earlier literature in small motion.
\end{abstract}

\begin{figure}[htb]
\begin{minipage}[b]{0.55\linewidth}
  \centering
  \centerline{\includegraphics[width=4.0cm]{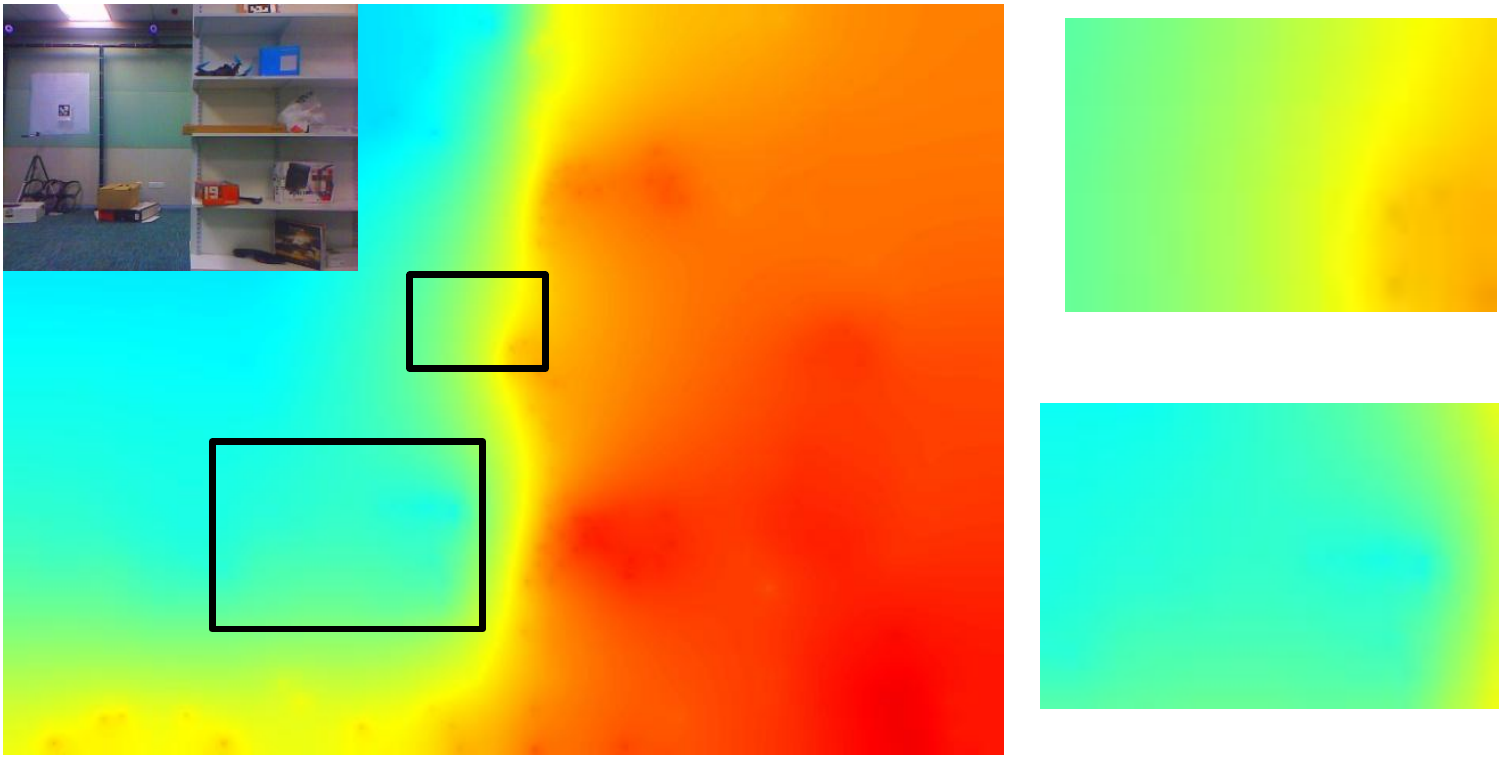}}
  \centerline{(a)}\medskip
\end{minipage}
\hfill
\begin{minipage}[b]{0.4\linewidth}
  \centering
  \centerline{\includegraphics[width=2.7cm]{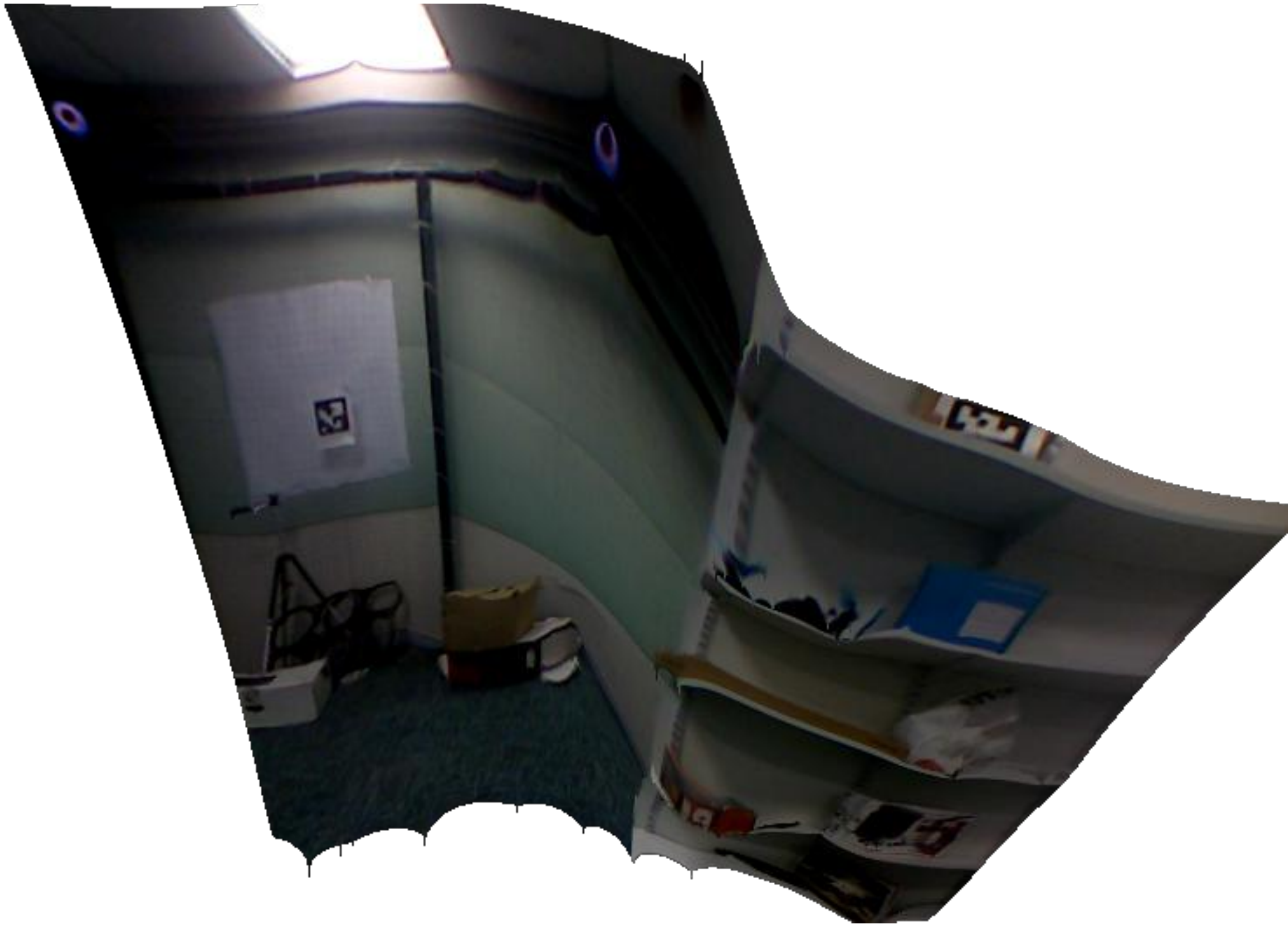}}
  \centerline{(b)}\medskip
\end{minipage}

\begin{minipage}[b]{0.55\linewidth}
  \centering
  \centerline{\includegraphics[width=4.0cm]{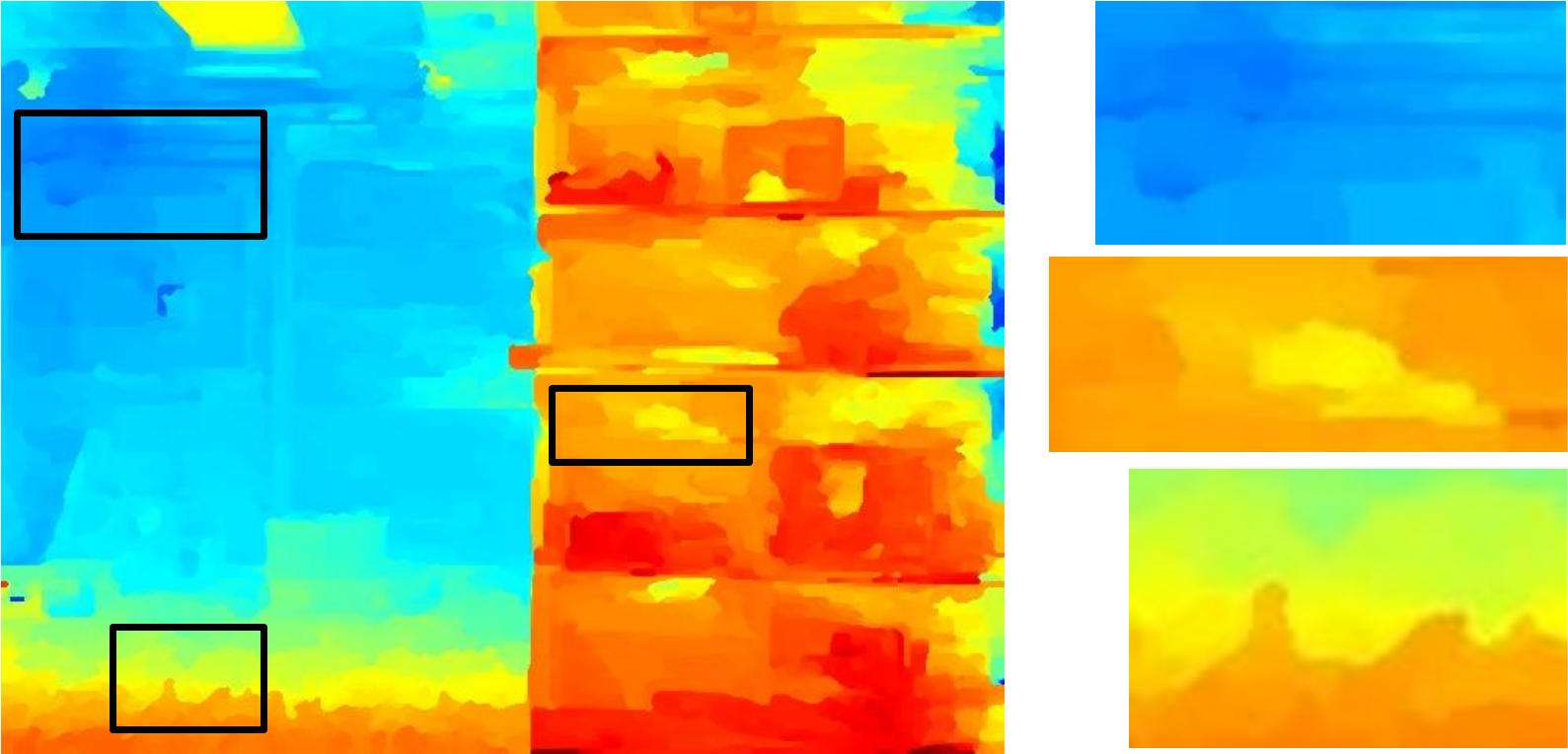}}
  \centerline{(c)}\medskip
\end{minipage}
\hfill
\begin{minipage}[b]{0.4\linewidth}
  \centering
  \centerline{\includegraphics[width=2.2cm]{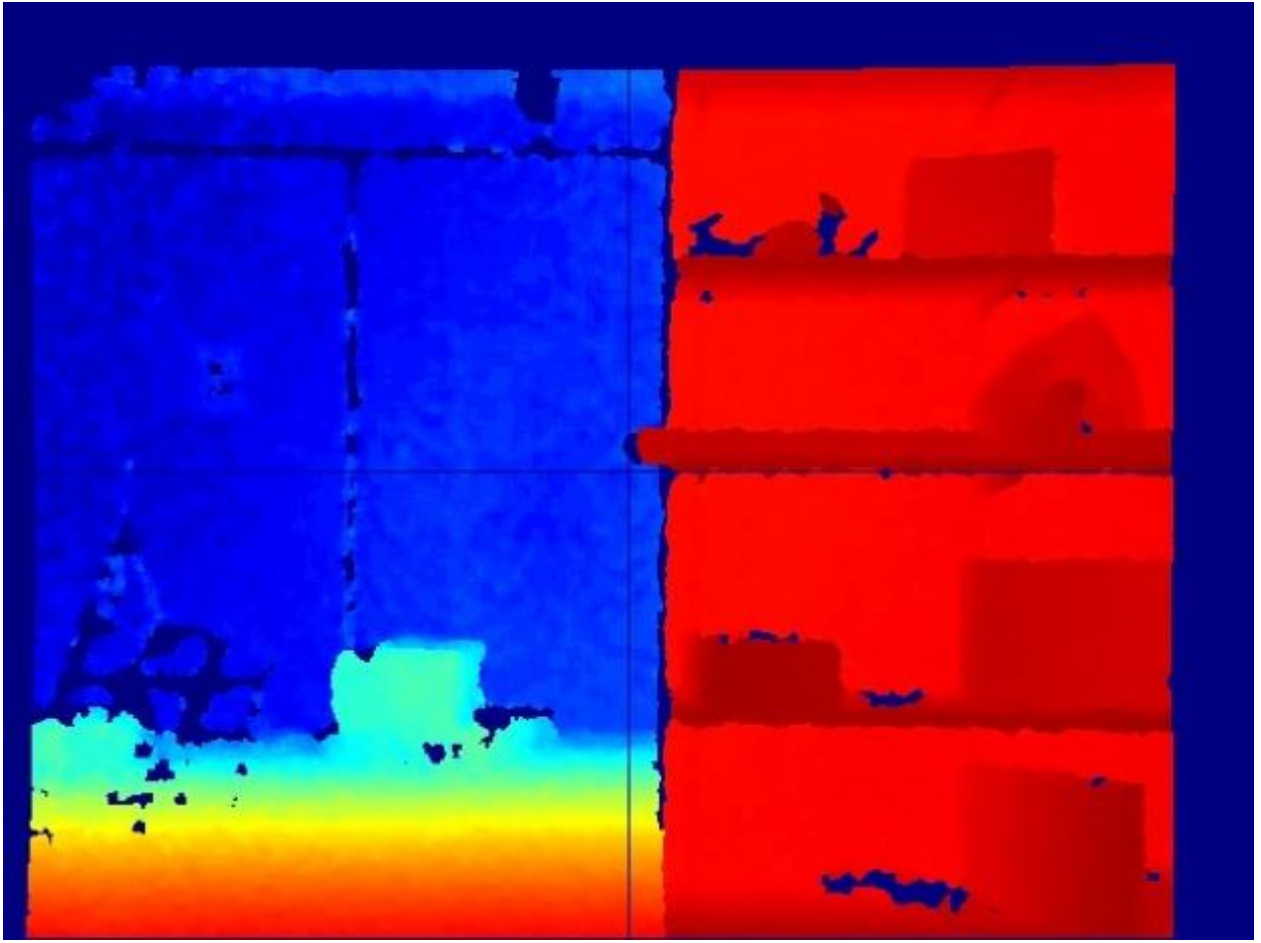}}
  \vspace{0.2cm}
  \centerline{(d)}\medskip
\end{minipage}

\begin{minipage}[b]{0.55\linewidth}
  \centering
  \centerline{\includegraphics[width=4.5cm]{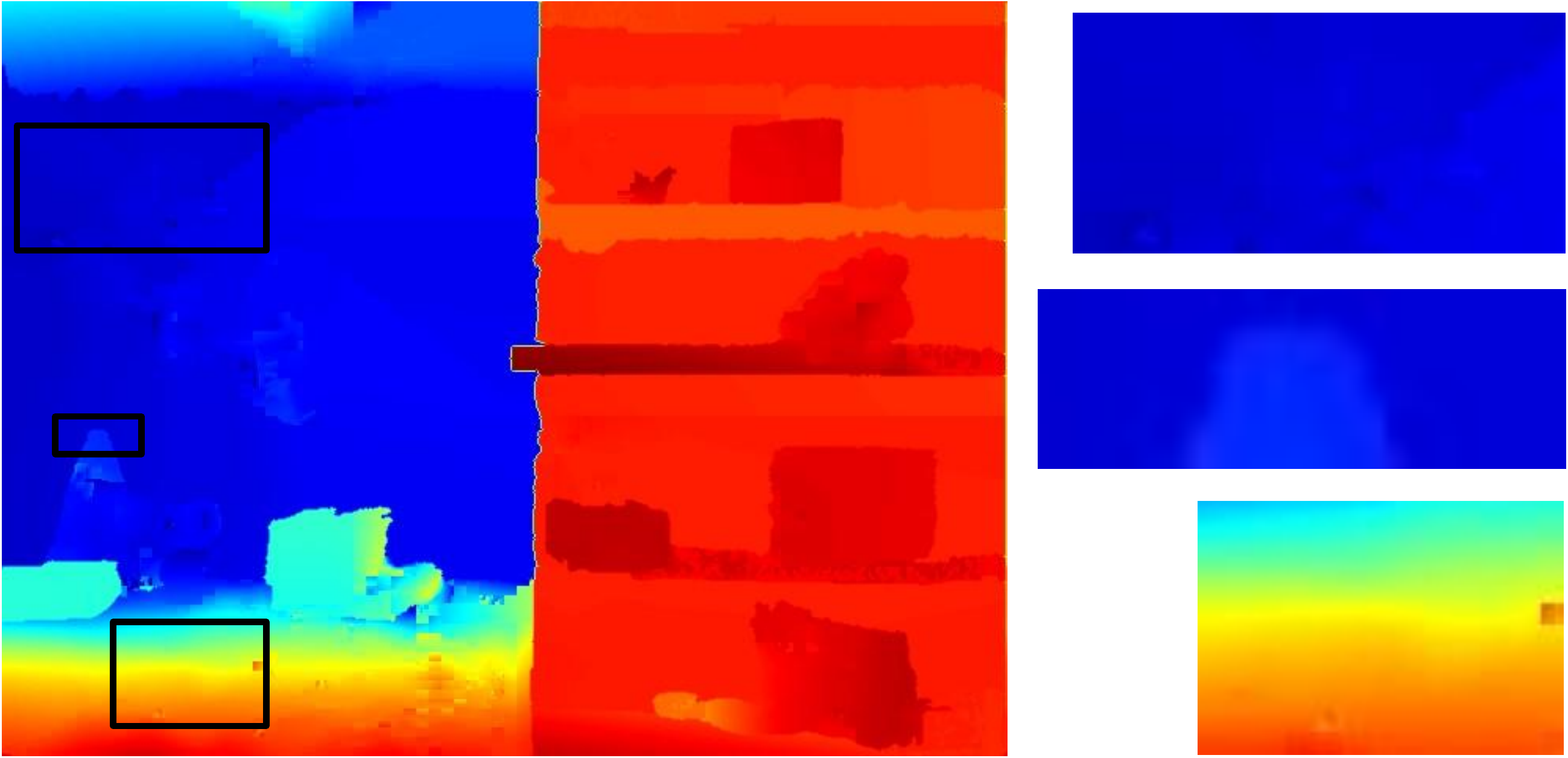}}
  \centerline{(e)}\medskip
\end{minipage}
\hfill
\begin{minipage}[b]{0.4\linewidth}
  \centering
  \centerline{\includegraphics[width=3.4cm]{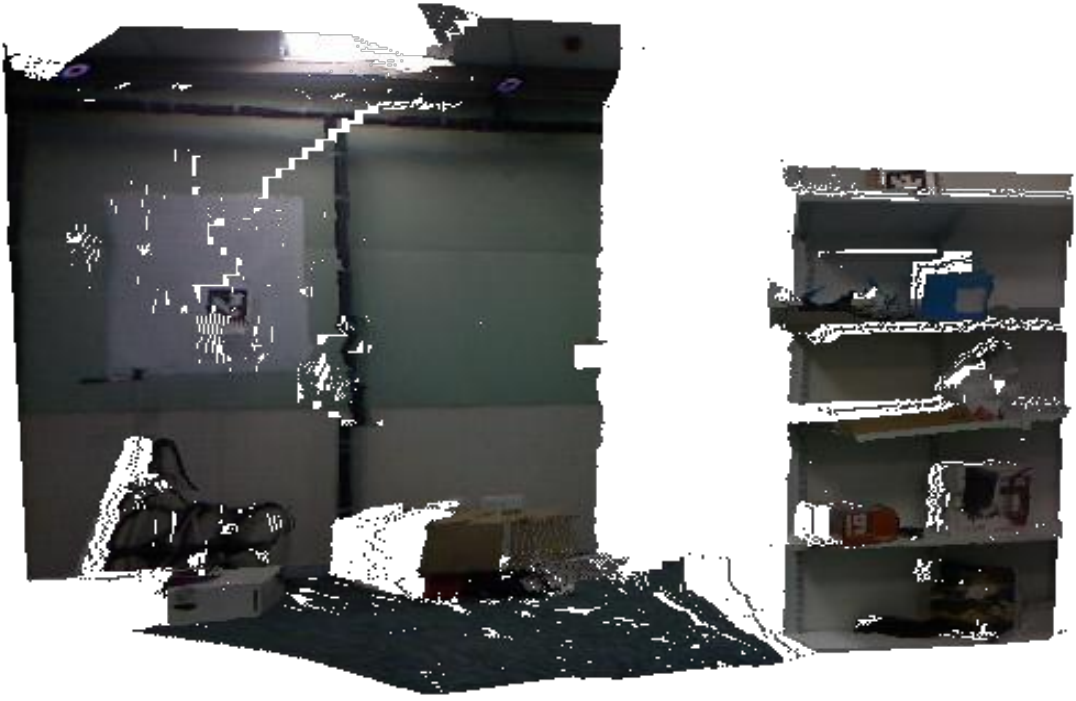}}
  \centerline{(f)}\medskip
\end{minipage}
\caption{{\bf Lab-room dataset.} Edge SLAM \cite{Maity_2017_ICCV_Workshops} reconstructs the sparse estimation. (a) Sample image and Depth map by \cite{imiccv15}. Depth estimation is very erroneous at object edge boundaries and obstacle on the floor is not detected. (b) 3D Point cloud by \cite{imiccv15}. Depth is continuous even at sudden depth changes. (c) Depth map by \cite{hacvpr16}. Depth on planar surfaces is erroneous. (d) Depth map by Kinect. (e) Our depth map. Depth estimation is better on planar surfaces as well as on the edges. (f) Our 3D point cloud where depth estimation on the entire scene is better compare to \cite{imiccv15} \& \cite{hacvpr16}.}
\label{fig:res1}
\end{figure}

\section{Introduction}
\label{sec:intro}

Autonomous navigation of Micro Aerial Vehicles (MAVs) in an outdoor cluttered environment is a well-researched topic in robotic research on last decade \cite{7139308, 6198031, langelaan2005towards, Selekwa:2008:RNV:1346348.1346426} but ignored for indoor navigation due to low-textured man-made environments. Autonomous navigation of MAVs requires robust estimation of a robot's pose along with dense depth for navigable space detection. Visual Simultaneous Localization and Mapping (vSLAM) address the problem of estimating camera pose along with 3D scene structure and it achieved significant improvement \cite{DBLP:journals/corr/CadenaCCLSN0L16}. Most of the existing vSLAMs produce a sparse 3D structure which is insufficient for navigable space detection but executing a real-time dense vSLAM is computationally very heavy. In contrast to a dense SLAM, dense depth map computation during hovering of a drone for locating free space is computationally less complex. After understanding the depth map drone can start moving using sparse vSLAM until it requires another understanding for new free space and hover again. While the drone is hovering, the baselines between consecutive images are very small, therefore, we need to estimate the depth map using the structure from Small Motion (SfSM) \cite{6909904} which has the advantages like better photometric constraints, rotation matrix simplification, etc. \cite{6909904} over regular motion. In this paper, we present a novel dense depth map estimation using SfSM at drone hovering.

\begin{figure}[htb]
\begin{minipage}[b]{0.325\linewidth}
  \centering
  \centerline{\includegraphics[width=2.6cm]{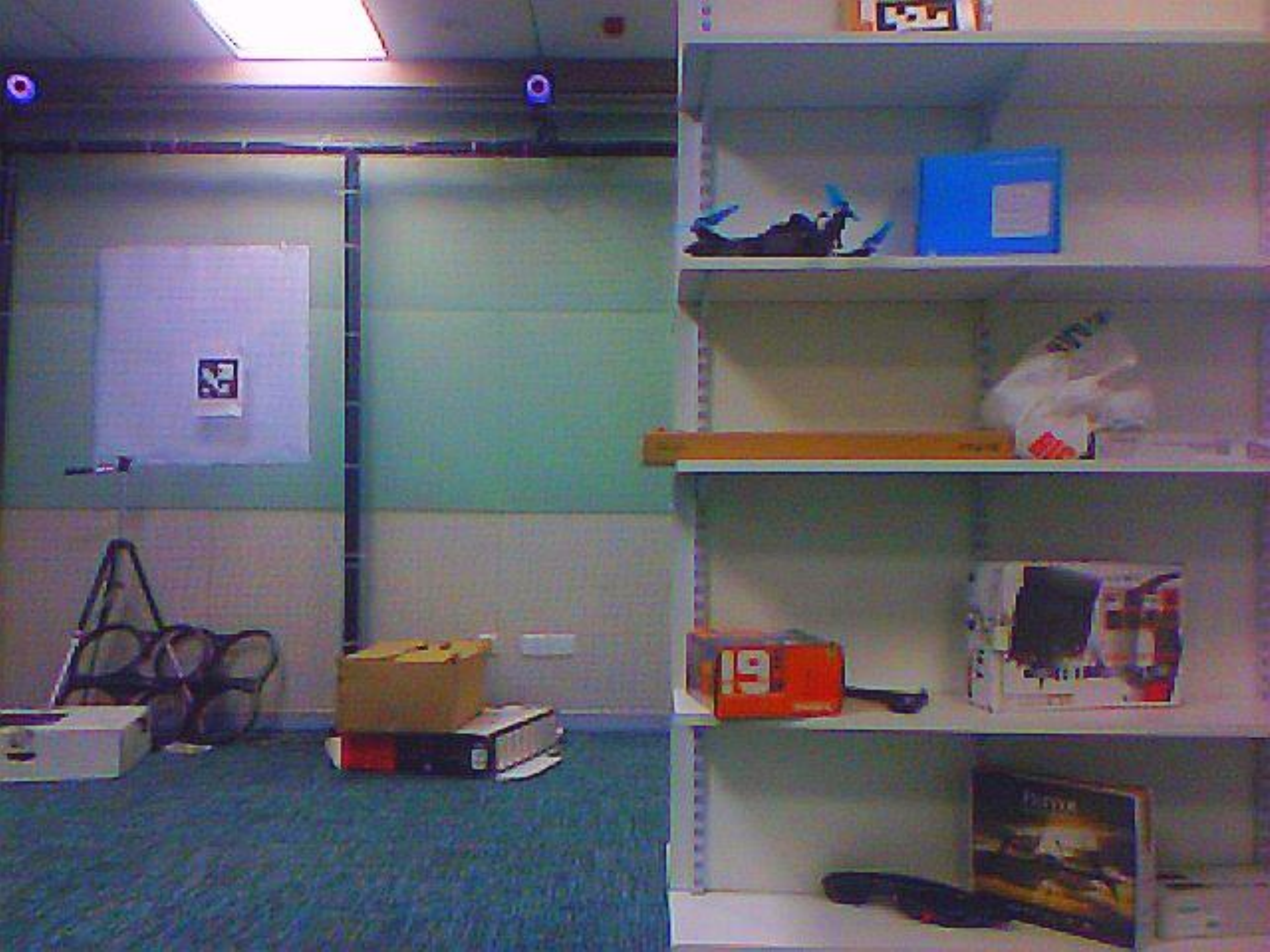}}
  \centerline{(a)}\medskip
\end{minipage}
\hfill
\begin{minipage}[b]{0.325\linewidth}
  \centering
  \centerline{\includegraphics[width=2.6cm]{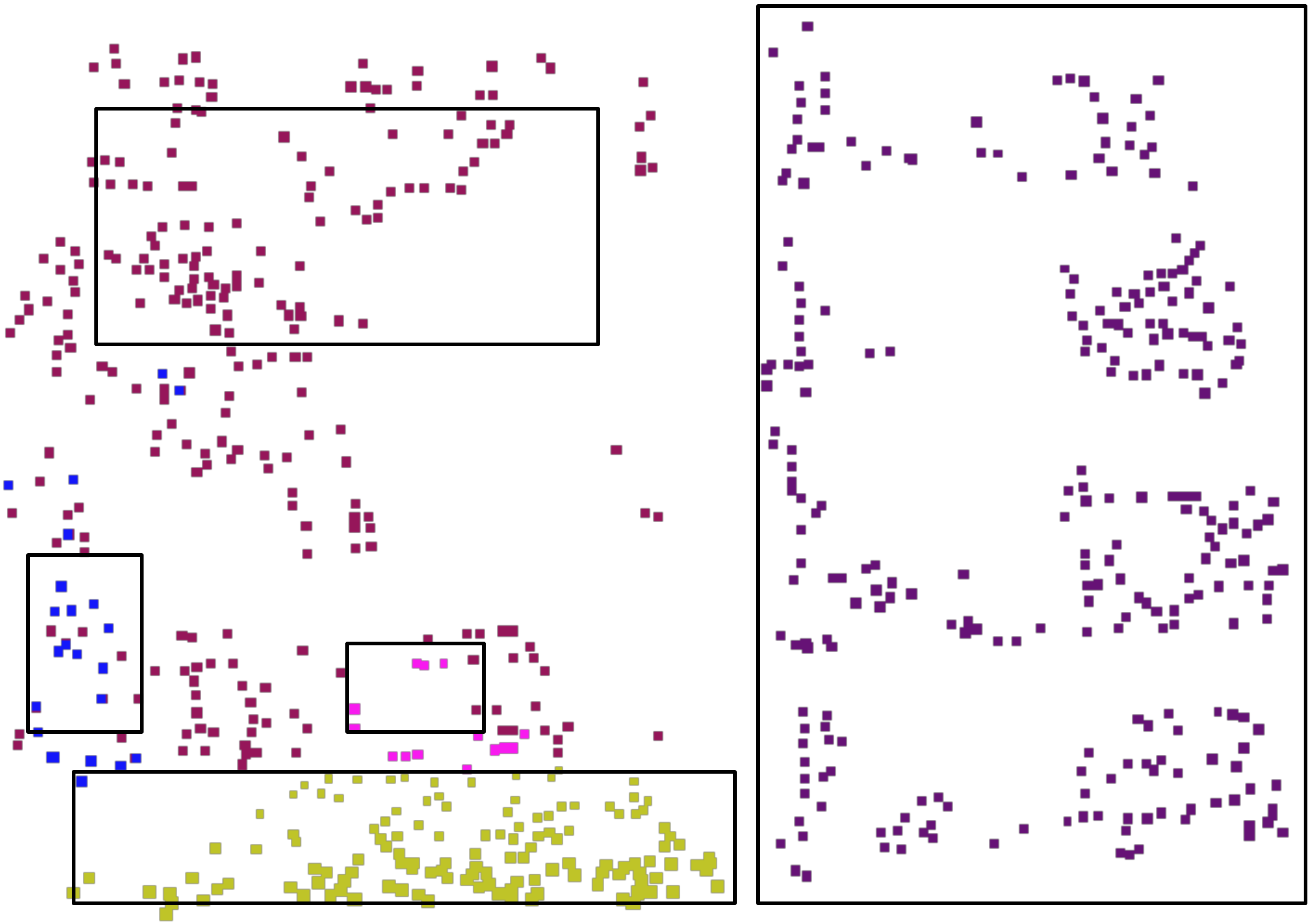}}
  \centerline{(b)}\medskip
\end{minipage}
\hfill
\begin{minipage}[b]{0.325\linewidth}
  \centering
  \centerline{\includegraphics[width=2.6cm]{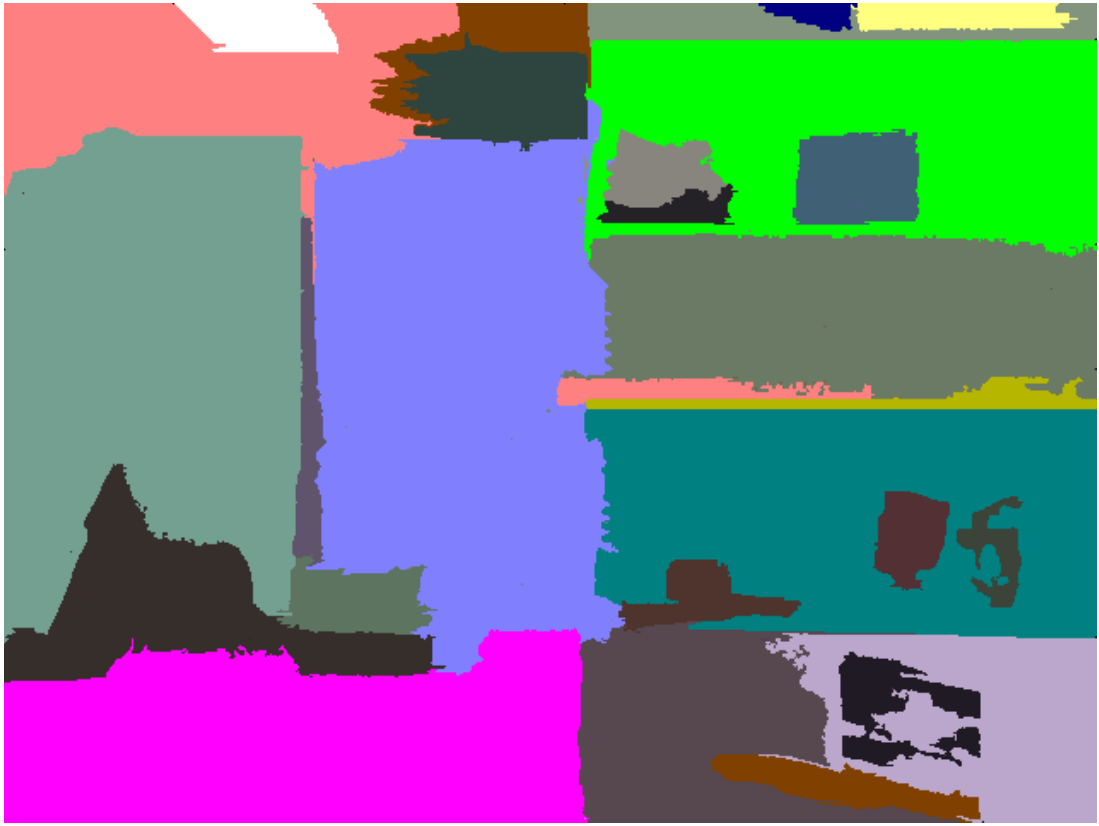}}
  \centerline{(c)}\medskip
\end{minipage}

\caption{(a) Reference image of {\bf Lab-room dataset}. (b) Segmented sparse 3D point cloud with 651 points. Segmented clusters are marked with bounded box for better understanding. {\bf Best viewed in colour.} (c) Segmented reference image.}
\label{fig:res2}
\end{figure}

Traditional feature based vSLAM approaches \cite{klein07parallel, mur2015orb, pumarola2017plslam, Maity_2017_ICCV_Workshops} produce erroneous metric depth when the baseline is less than $\sim$8 mm and viewing angle of a 3D point is less than 0.2$^{\circ}$ \cite{6909904}. Despite several works on SfSM \cite{6909904, micro-baseline-stereo, imicip15, imiccv15, hacvpr16, Javidnia_2017_ICCV_Workshops, Ham}, there exists inaccuracy in the estimation of dense depth for navigable space detection on the fly. Im \textit{et al.} \cite{imicip15} present a SfSM system that uses a plane fitting using colour and geometry consistency. Im \textit{et al.} propose another SfSM algorithm \cite{imiccv15} using plane fitting through a quadratic energy function consists of sparse data points, colour and geometric smoothness. Both the systems \cite{imicip15, imiccv15} use sparse 3D points for geometric validation, which is erroneous in low-textured indoor environments due to less number of features. Both the systems \cite{imicip15, imiccv15} fit a continuous plane in the entire visible 3D space which fails to estimate sudden depth variation. Fig.~\ref{fig:res1}(a), (b) show such erroneous depth estimation by \cite{imiccv15}. Recently, Ha \textit{et al.} present a SfSM system \cite{hacvpr16} that uses a plane sweep technique which produces discrete dense depth and fails to estimate reliable depth for regular planar surfaces along with edge boundaries. The plane sweep runs on image pixels without considering the sparse point position which allows a pixel to attain a particular depth without any neighbouring constrain and estimate noisy depth on a planar surface as shown in Fig.~\ref{fig:res1}(c). The method is not real-time (takes around 10 minutes). Javidnia and Corcoran present a ORB feature \cite{Rublee:2011:OEA:2355573.2356268} based SfSM system \cite{Javidnia_2017_ICCV_Workshops} where plane sweeping technique generates dense depth directly from ORB feature matches. The ORB feature highly depends on texture present on the scene, therefore the depth map from \cite{Javidnia_2017_ICCV_Workshops} is also erroneous in the low-textured environment. The system running time is about 8 minutes claimed by the authors. So there exists no system which produces reliable dense depth for free space estimation in indoor.

In this work, we target to obtain an accurate dense depth map for navigable space detection by a drone in indoor environments using small motion configuration \cite{6909904}. We use Edge SLAM \cite{Maity_2017_ICCV_Workshops}, the most resilient SLAM in low-textured environment, for estimating camera poses and initial sparse 3D points while the drone is in normal motion and allow the drone to hover and continue estimating the small motion camera poses using resectioning \cite{Epnp09}. We start the dense depth estimation as soon as we get at least 20 estimated images from Edge SLAM with small motion where the first image in small motion is the reference image. We introduce a segmentation step at the beginning in contrast with previous approaches \cite{imicip15, imiccv15, hacvpr16, Javidnia_2017_ICCV_Workshops}, to segment the reference image into smaller regions to avoid erroneous depth estimation through continuous plane fitting. Segmentation was never been used in the literature of SfSM and this is a contribution in this work. Depth propagation runs on every image segment using a novel patch-based (a minimal segmented 2D area which can be considered as a planar surface in 3D) plane fitting approach through minimizing a cost function consisting of sparse 3D points, intensity and gradients of pixels and co-planarity with neighbouring patches. We design the patch-based cost function for each image segment independently that produces a better depth estimation compared with earlier approaches (e.g.\ Fig.~\ref{fig:res1}), which is the major contribution of this work. We attach a weight with every patch based on its depth initialization and further add that patch for optimization when the weight is below to certain threshold. This novel idea of weighted patch-based optimization is computationally lightweight which makes the method suitable for real-time systems like drone navigation and we achieve $\sim$14 seconds in execution time to estimate dense depth using 20 images.

Section~\ref{sec:Method} illustrates the insights of our proposed method. Section~\ref{ssec:result} provides the experimental results. Section~\ref{ssec:conclusion} concludes this paper.




\section{Proposed Methodology}
\label{sec:Method}
\subsection{Segmentation}
\label{ssec:segments}
We segment the sparse point cloud using a combination of nearest neighbour clustering and colour similarity between 3D points based on \cite{RusuDoctoralDissertation}. We use a Kd-tree representation of input sparse point cloud. The clustering algorithm generates clusters of points having close location proximity (lower than an adaptive proximity threshold based on camera locations) with similar colour. Fig.~\ref{fig:res2}(b) show the segmented point cloud of the Lab-room dataset where the floor, walls, objects belongs to separate segments.

We apply a non-linear bilateral filter \cite{Tomasi:1998:BFG:938978.939190} to de-noise the reference image. The basic idea of bilateral filtering is to do an additional filtering in range domain along with traditionally filtering in the spatial domain. We use a bilateral filter to preserve the edge properties of the image as it combines the pixel intensities based on their geometric as well as photometric closeness. Further, we segment the filtered image using a colour based region growing segmentation based on \cite{Tremeau:1997:RGM:1746432.1746438}. We further use the segmented sparse 3D point clusters to merge 2D segments. 3D points present in a cluster should lie on the same image segment. We project the 3D points in a cluster on the image plane and merge the smaller image segments based on spatial information (e.g.\ Fig.~\ref{fig:res2}(c)). Our depth propagation is erroneous only if the segmentation fails both in 2D and 3D. But such cases occur only when an entire segment is of similar colour and no sparse point present to differentiate abrupt depth change in 3D but the occurrence of such situations is very infrequent.

\subsection{Depth Propagation}
\label{ssec:dense}

Depth propagation method estimates the dense depth for every image segment in patch-based. An image segment becomes a set of patches $\omega$ and another set of neighbouring patches $\mathcal{N_{\omega}}\ \forall\ p\in\omega$. Every patch corresponds to a planar 3D surface characterized as ${\zeta}_p = \{\nu_p, \vec{n}_p\}$. $\nu_p$ represents a point on ${\zeta}_p$ and $\vec{n}_p$ represents the normal vector of ${\zeta}_p$ from $\nu_p$. $\theta$ is the set of pixels in patch $p$, ${\varphi}_{p}$ represents the set of sparse 3D points whose projection lies within the patch $p$ boundary and $\eta = |{\varphi}_{p}|$. We initialize all the patches $p\in\omega$ with a planar surface in the most feasible way. We start with the patch that has largest $\eta$. Initialize it by fitting a 3D planar surface using the points in ${\varphi}_{p}$ having reprojection error $< 0.1$ pixel. Subsequently, continue the initialization for neighbouring patches where $\eta > 0$ using a similar plane fitting technique. There still exist uninitialized patches where $\eta = 0$ and we initialize them by connecting the 3D surfaces of already initialized neighbouring patches. There may exist image segments where $\eta = 0\ \forall\ p\in\omega$ and initialization fails. We calculate the gradient of every patch by adding intensity gradients of all pixels belong to the patch $p$. We use the plane sweep approach as proposed in \cite{hacvpr16} only for the patch with largest gradient. Plane sweep runs a virtual plane perpendicular to the viewing direction from maximum to minimum depths obtained from sparse point cloud. The particular depth with minimum photometric error initializes the dense depth for that patch. Subsequently, continue the initialization for neighbouring patches using continuous plane fitting. We formulate the cost $C(\zeta)$ as in Eq.~\ref{equ:cost} and minimize it through the parameters $\nu_p$ and $\vec{n}_p$.

\begin{equation}
C(\zeta) = \sum_{p\in\omega} {\lambda}_{p} C_{p}({\Psi}_{p}^{D} + {\Psi}_{p}^{I} + {\lambda}_{g}{\Psi}_{p}^{G}) \\ 
+ {\tau} \sum_{(p,q)\in\mathcal{N_{\omega}}} {\lambda}_{pq} C_{p} {\Psi}_{pq}^{C}
\label{equ:cost}
\end{equation}

where ${\lambda}_{p}$ and ${\lambda}_{pq}$ represent adaptive normalizing weights. $C_{p}$ is a confidence weight for every patch $p$. ${\lambda}_{g}$ represents the weight for ${\Psi}_{p}^{G}$ and $\tau$ represents a balancing factor between the data term and the regularization term. 

\begin{equation}
{\lambda}_{p} = {\sigma}{\beta}_p
\end{equation}

where ${\sigma}{\beta}_p$ represents the variation in projected area of 2D patches using the points in $\theta$ and all views $\upsilon \in V$.

\begin{equation}
{\lambda}_{pq} = \frac{1}{\xi} \sum_{x \in \xi} {\delta(x)}
\end{equation}

where $\xi$ represents the set of edge pixels between $p$ and $q$. $\delta(x)$ is the intensity gradient of the pixels $x \in \xi$. We design $C_{p}$ as in Eq.~\ref{euq:choose}, which speeds up the optimization by providing lower weight to stable patches.

\begin{figure}[htb]
\begin{minipage}[b]{0.32\linewidth}
  \centering
  \centerline{\includegraphics[width=2.7cm]{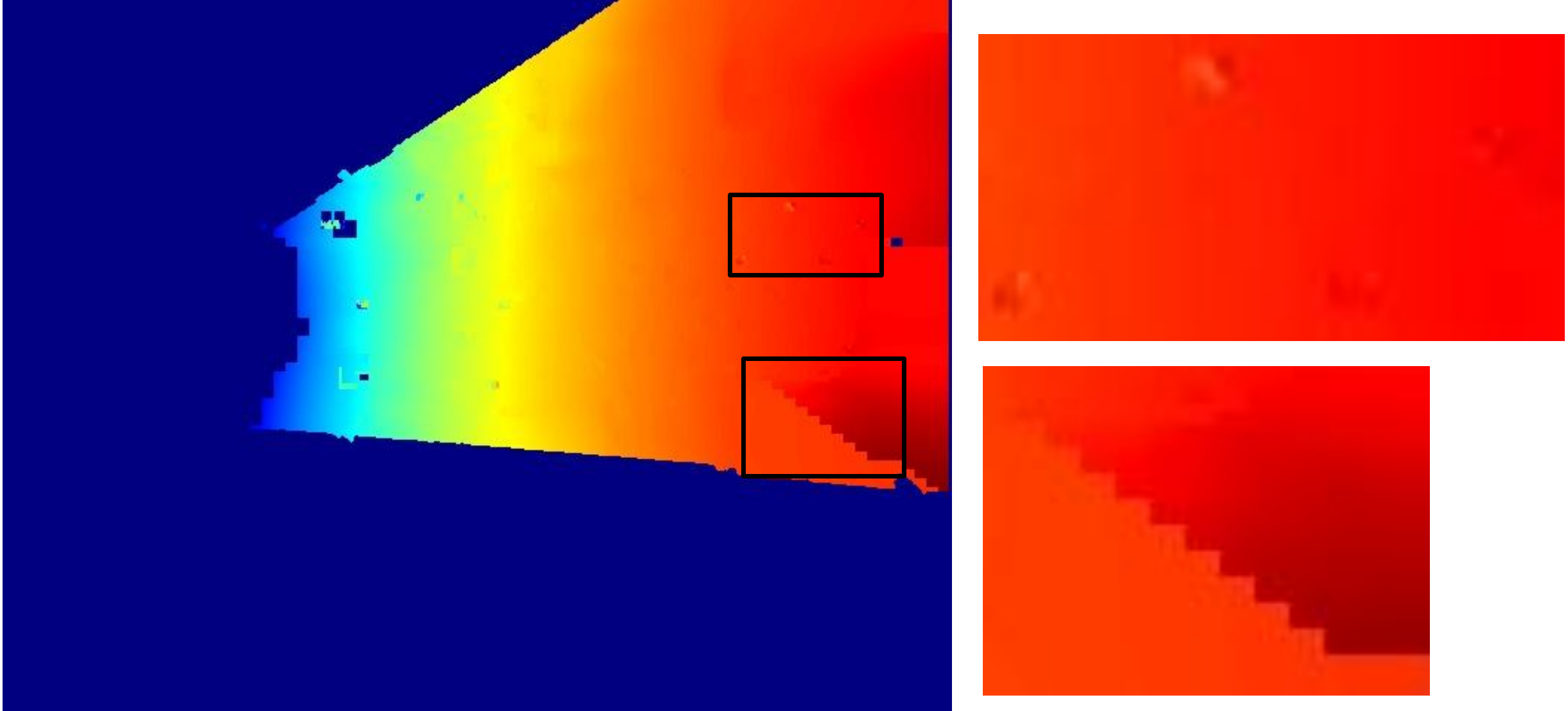}}
  \centerline{(a)}\medskip
\end{minipage}
\hfill\begin{minipage}[b]{0.33\linewidth}
  \centering
  \centerline{\includegraphics[width=2.7cm]{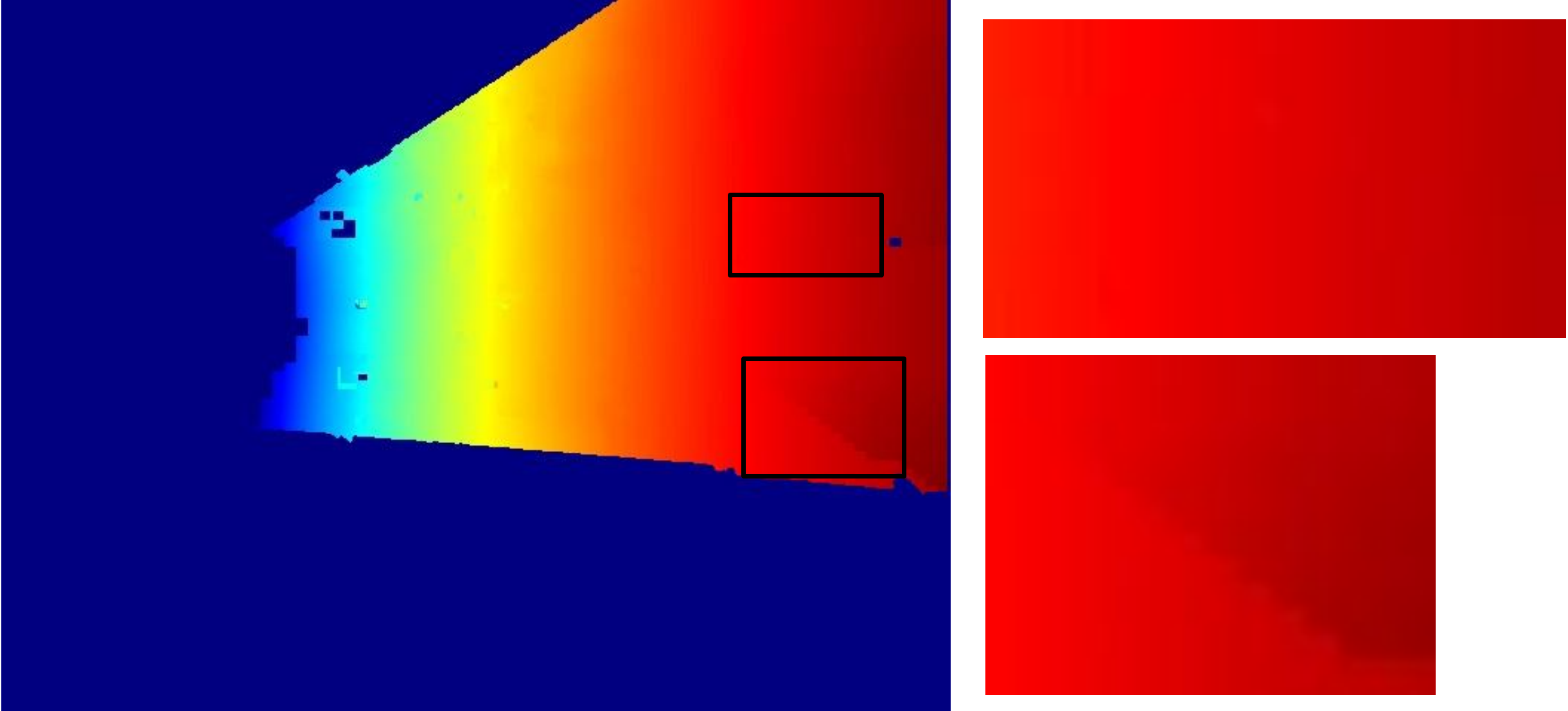}}
  \centerline{(b)}\medskip
\end{minipage}
\hfill
\begin{minipage}[b]{0.33\linewidth}
  \centering
  \centerline{\includegraphics[width=2.7cm]{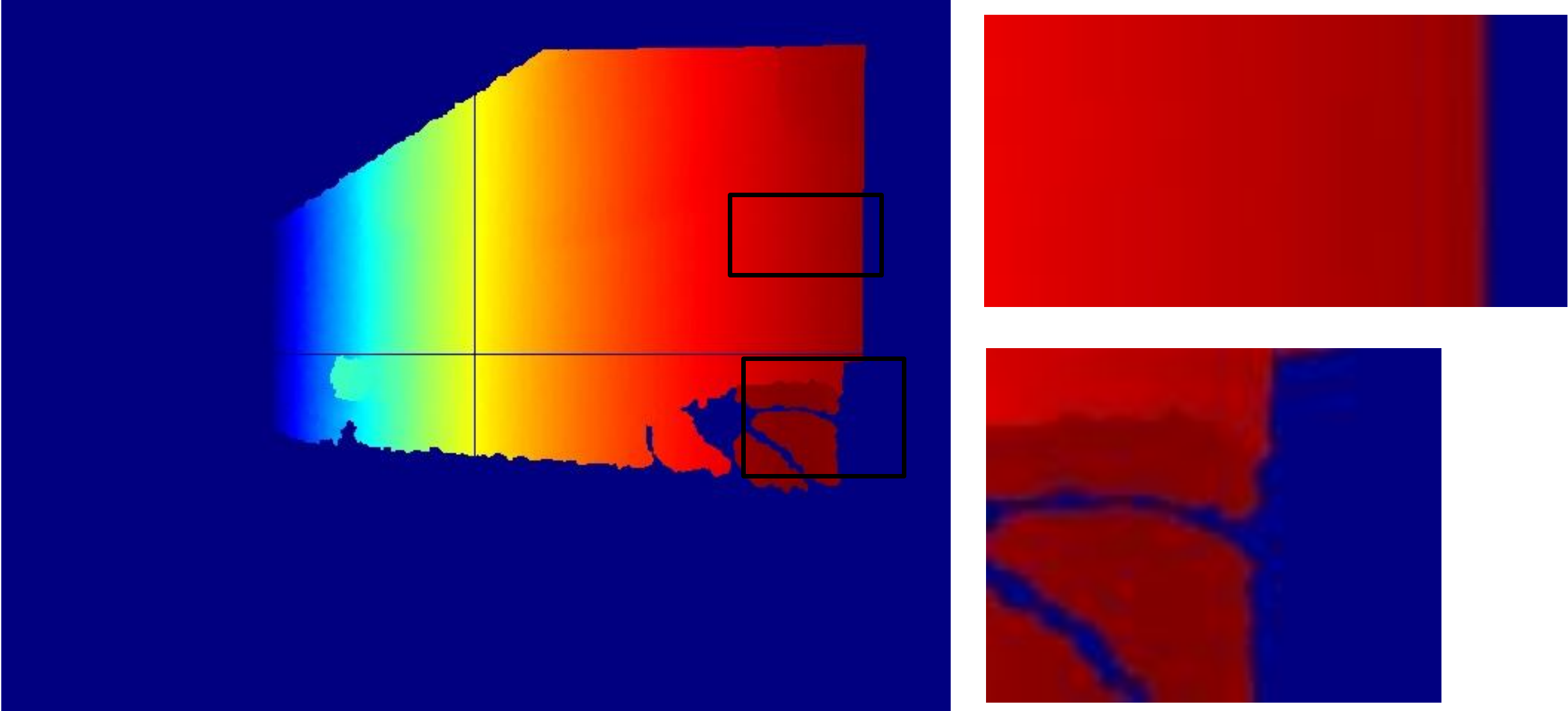}}
  \centerline{(c)}\medskip
\end{minipage}
\caption{(a) Depth map of a sample image segment with our weighted patch-based optimization. (b) Depth map of the segment with full optimization. (c) Depth map using Kinect.}
\label{fig:res3}
\end{figure}

\begin{equation}
C_{p} = 
\begin{cases}
	0 & \text{: if ($\eta * area({\varphi}_{p})$) $\geq$ $\Delta$} \\
	\frac{1}{\eta * area({\varphi}_{p})} & \text{: $\eta$ $\geq$ $0$} \\
	1 & \text{: Otherwise}
\end{cases}
\label{euq:choose}
\end{equation}

where $area({\varphi}_{p})$ represents the 2D projected area of points in ${\varphi}_{p}$ and $\Delta$ is a given threshold. A patch with $C_{p} = 0$ is considered as stable because $(\eta * area({\varphi}_{p})) \geq \Delta$ and therefore excluded in optimization. Fig.~\ref{fig:res3} shows a comparison result where $C_{p} = 0$ for 258 patches, $C_{p} < 1$ for 636 patches and $C_{p} = 1$ for 1865 patches in the image segment. Our weighted patch-based optimization performs $\sim$9.7 times faster compared to full optimization. The full optimization reduces some artefacts (shown in inset) with a smoothness on the surface, but this little improvement has almost no impact on free space calculation.

{\bf Sparse 3D point consistency:} We formulate the sparse point consistency ${\Psi}_{p}^{D}$ as:

\begin{equation}
{\Psi}_{p}^{D} = \sum_{i\in{\varphi}_{p}} {\frac{1}{\partial}(\|X_i - X_{i {\zeta}_p}\|)}
\end{equation}

where $\partial$ denotes the average reprojection error for the point $i\in{\varphi}_{p}$. $X_i$ \& $X_{i{\zeta}_p}$ denote the sparse depth and the depth on the surface ${\zeta}_p$ for the point $i$. ${\Psi}_{p}^{D}$ provides lower weight to the sparse 3D points having high reprojection error.

\begin{figure*}[htb]
\begin{minipage}[b]{0.14\linewidth}
  \centering
  \centerline{\includegraphics[width=2cm]{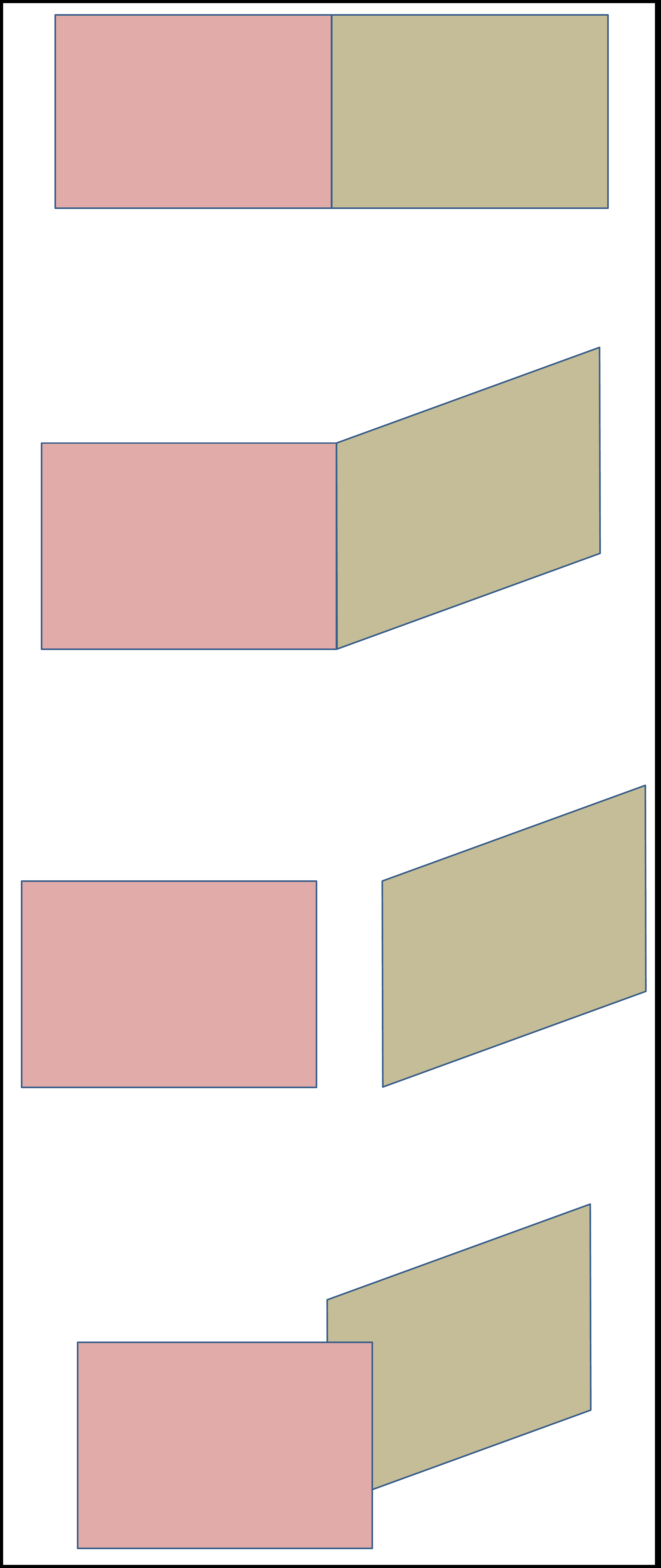}}
  \vspace{0.4cm}
  \centerline{(a)}\medskip
\end{minipage}
\hfill
\begin{minipage}[b]{0.55\linewidth}
  \centering
  \centerline{\includegraphics[width=10cm]{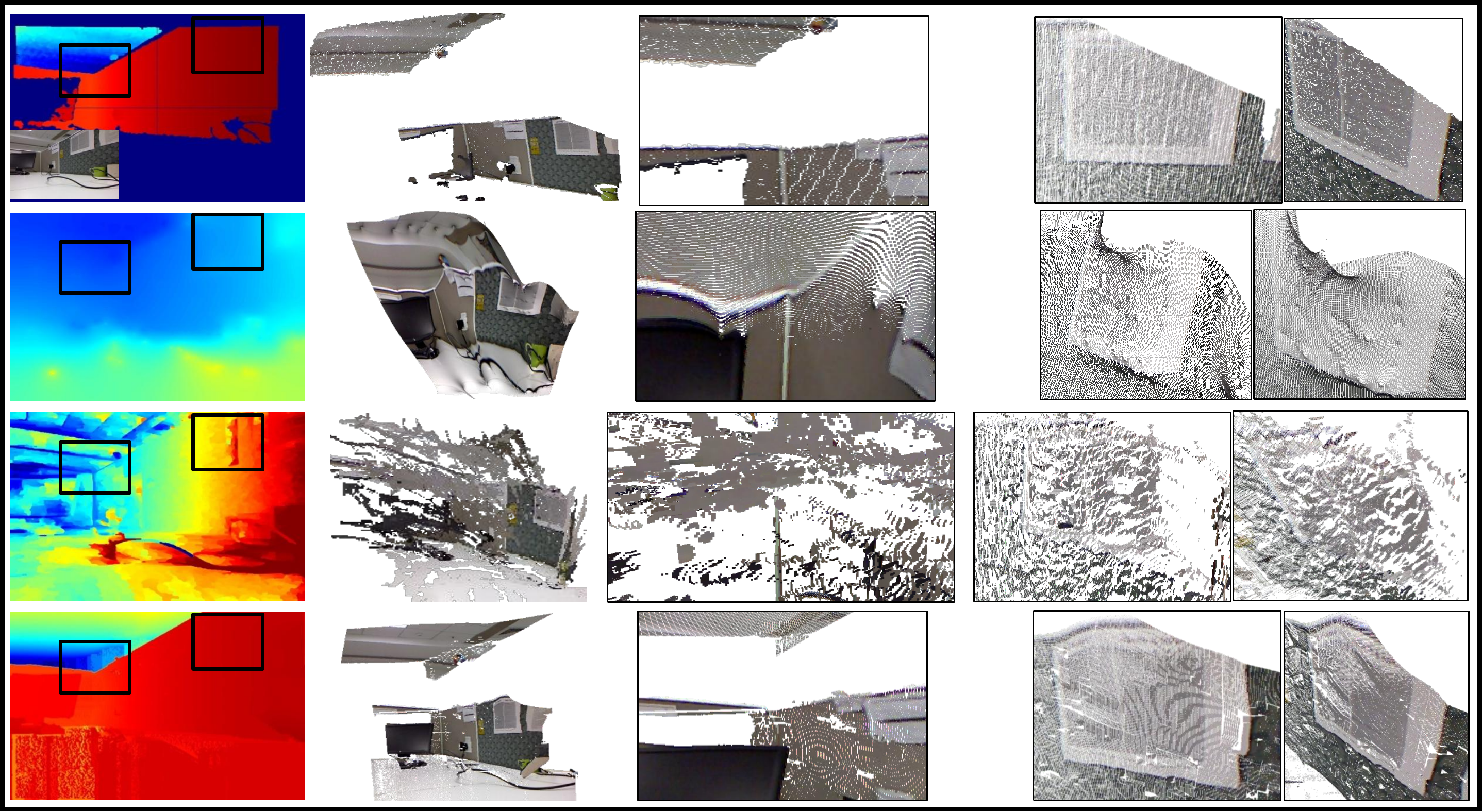}}
  \centerline{(b)}\medskip
\end{minipage}
\hfill
\begin{minipage}[b]{0.23\linewidth}
  \centering
  \centerline{\includegraphics[width=3.8cm]{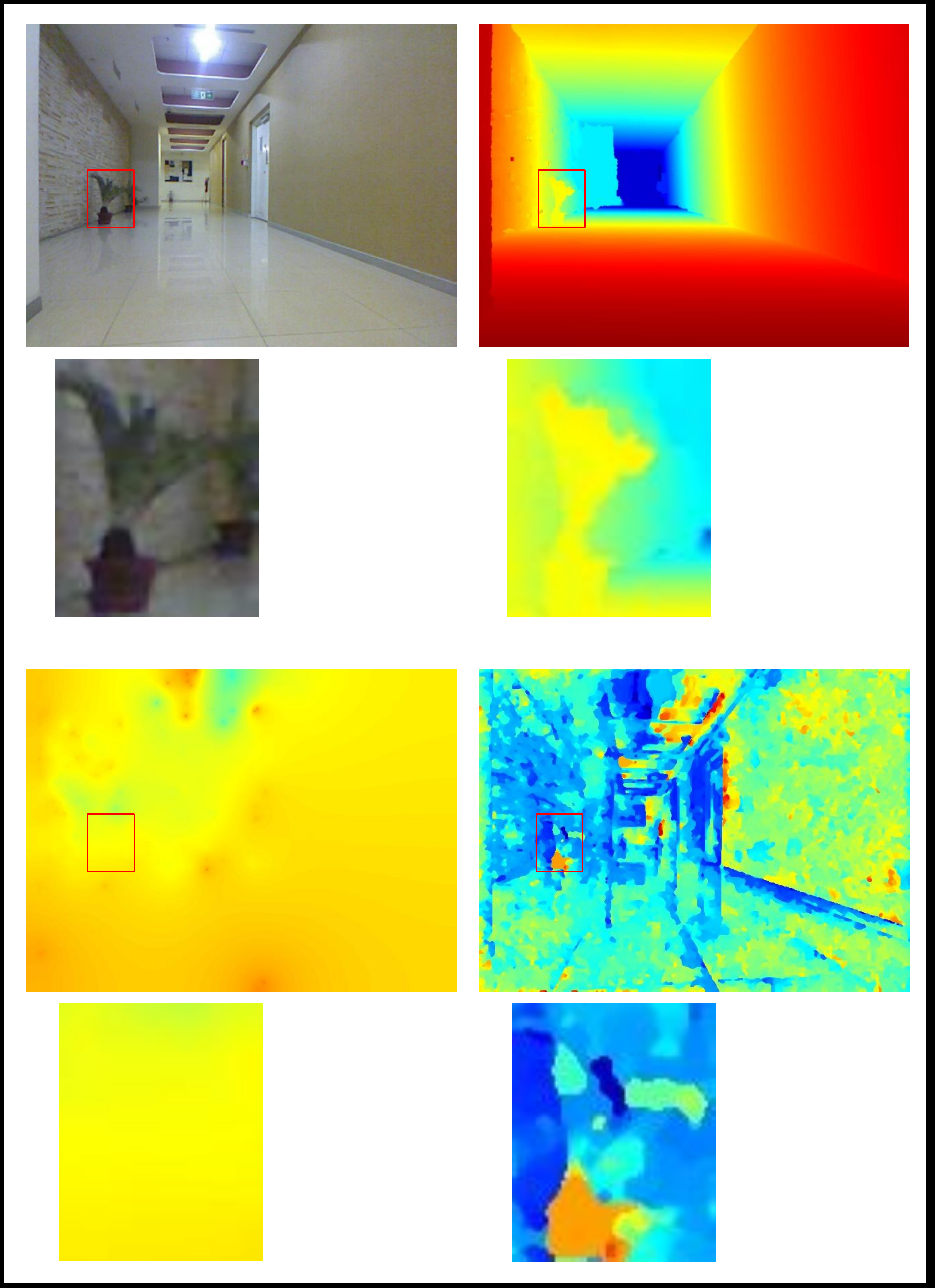}}
  \vspace{0.13cm}
  \centerline{(c)}\medskip
\end{minipage}

\caption{(a) Four types of neighbouring plane association that are considered in optimization. (b) {\bf Office-Space dataset.} 715 sparse points. A comparative result between Kinect (1st row), proposed method by \cite{imiccv15} (2nd row), proposed method by \cite{hacvpr16} (3rd row) and our method (4th row). We present depth maps (1st column), point cloud (2nd column), enlarged areas (3rd and 4th columns). (c) {\bf Corridor dataset.} 557 sparse points. 1st Row: Sample image of the dataset and Depth map using our proposed method. 3rd Row: Depth map by \cite{imiccv15} and depth map by \cite{hacvpr16}. 2nd \& 4th Row: Enlarged inset images.}
\label{fig:res4}
\end{figure*}

{\bf Photo-consistency:} The term ${\Psi}_{p}^{I}$ in Eq.~\ref{equ:cost} represents the cost of intensity matching between a patch $p$ and it's warped patches $\pi_p(\upsilon)$ in other views $\upsilon \in V$ using plane-induced homography \cite{Hartley2004} in small baseline. $\pi_p(\upsilon)$ is designed as:

\begin{equation}
\pi_p(\upsilon) = {\lambda}_{\upsilon} H(p, \{\nu_p, \vec{n}_p\})
\end{equation}

where $H$ is the homography matrix \cite{Hartley2004} of surface ${\zeta}_p$ on the view $\upsilon$, ${\lambda}_{\upsilon}$ is the weight for view $\upsilon$ which provides more priority to nearer frames because intensity variation is less sensitive to closer views with depth variation. We design the intensity cost as:

\begin{equation}
{\Psi}_{p}^{I} = \sum_{p \in \omega} \sum_{x \in \theta} {Var(I_{1{\zeta}_{p}}, \cdots, I_{\upsilon{\zeta}_{p}})}
\label{equ:var}
\end{equation}

where $Var(\cdotp)$ is the variance function and $I_{i{\zeta}_{p}}$ represents the intensity of the pixel $x\in \theta$ in view $\upsilon$ projected from surface ${\zeta}_{p}$. We define a more robust variance calculation by providing higher priority to images nearer to the reference image. The term ${\Psi}_{p}^{G}$ in Eq.~\ref{equ:cost} represents the cost of gradient matching which follows the same variance formula as described in Eq.~\ref{equ:var} with the gradient of intensity at the place of intensity.

{\bf Regularization:} The pairwise regularization works on the assumption of connectivity with neighbouring patches at edge boundaries. We consider five types of priority based patch pair configuration to cater all possible occupancy conditions. Fig.~\ref{fig:res4}(a) shows four such pairs refer in Eq.~\ref{equ:switch}.

\begin{equation}
{\Psi}_{pq}^{C} = 
\begin{cases}
	0 & \text{:for connected surfaces (1st \& 2nd rows)} \\
	{\varrho}_{1} & \text{:for disconnected surfaces (3rd row)} \\
	{\varrho}_{2} & \text{:for ocluded surfaces (4th row)} \\
	{\varrho}_{3} & \text{:otherwise}
\end{cases}
\label{equ:switch}
\end{equation}

where $0 < {\varrho}_{1} < {\varrho}_{2} < {\varrho}_{3}$

\section{Experimental Results}
\label{ssec:result}

We use an Intel i7-8700 (6 cores @3.7-4.7GHz) with 16Gb RAM, for implementation and a Bebop quadcopter from Parrot Corporation for data acquisition. We consider registered Kinect point cloud as ground truth as shown in Fig.~\ref{fig:res1}. We perform all our experiments on 640x480 VGA resolution. All experiments use parameters: $\Delta = 7$, ${\lambda}_{g} = 3$, $\tau = 1.7$, ${\varrho}_{1} = 0.6$, ${\varrho}_{2} = 3.5$, ${\varrho}_{3} = 20$.

We present two indoor datasets suitable for drone navigation to evaluate the performance of our proposed method in comparison with earlier methods. Sparse reconstruction in all
the cases are executed using Edge SLAM \cite{Maity_2017_ICCV_Workshops}. Fig.~\ref{fig:res4}(b) shows {\bf Office-Space} dataset, an indoor office premise with smooth
depth variation along the wall and a sudden depth changes at wall boundary. We compare our result with Kinect and the results obtained using \cite{imiccv15} and \cite{hacvpr16}. We show a region with sudden depth variation (\textit{3rd column}) and another region with a smooth planar surface (\textit{4th column}). Both the earlier methods \cite{imiccv15, hacvpr16} fail to estimate sudden depth changes at object boundary and erroneous depth estimation on the planar surface. 

due to continuous plane fitting by \cite{imiccv15} and erroneous plane sweep
by \cite{hacvpr16} as already explained in Sec.~\ref{sec:intro}. The mean error in depth
estimation using our method against Kinect point cloud is 0.2352 meter whereas method by \cite{hacvpr16} producing 2.11575 meters errors. Fig.~\ref{fig:res4}(c) represents a {\bf Corridor} dataset, a corridor with an object on the ground. Our estimation shows better
accuracy in estimating depth variation on the wall as well as on the ground compared to \cite{imiccv15, hacvpr16} for the similar reasons. Our dense depth estimation has least artefacts and more realistic for free space understanding in both cases.

{\bf Execution Time:} Timing is a vital factor for any method to run on the drone. Our method achieves a running time of $\sim$14 sec using 20 images on our unoptimized implementation.

\section{Conclusion}
\label{ssec:conclusion}

We have demonstrated an approach to estimate indoor dense depth map on drone hovering using camera calibration and sparse point cloud from Edge SLAM. We have introduced a novel segmentation procedure at beginning to segment the reference image in objects and estimate depth independently for each segment which improves the accuracy in depth estimation. We have proposed a novel patch-based plane fitting approach for depth estimation through minimizing a cost function which consists of sparse points, photo consistency and regularization terms. Our method attended a significant improvement in term of accuracy and feasibility for any real-time platform like drones.

\bibliographystyle{IEEEbib}
\bibliography{smallMotionDense}

\end{document}